\def\year{2019}\relax
\documentclass[letterpaper]{article} 

\usepackage{aaai19}  
\usepackage{times}  
\usepackage{helvet}  
\usepackage{courier}  
\usepackage{url}  
\usepackage{graphicx}  
\usepackage[usenames,dvipsnames]{color}
\usepackage{tikz}
\usepackage{amsmath}
\usepackage{amsfonts}
\usepackage{amssymb}
\usepackage{bussproofs}
\usepackage[ruled,vlined]{algorithm2e}
\usepackage[normalem]{ulem}

\newenvironment{bprooftree}  
  {\leavevmode\hbox\bgroup\bottomAlignProof}
  {\DisplayProof\egroup}

\SetKwComment{Comment}{\# }{}  

\DeclareMathOperator*{\argmax}{arg\,max}
\DeclareMathOperator*{\argmin}{arg\,min}
\DeclareMathOperator*{\maximize}{maximize}
\DeclareMathOperator{\E}{\mathbf{E}}

\begin{document}
%
\title{Automated Theorem Proving in Intuitionistic Propositional Logic\\ by Deep Reinforcement Learning}
\author{Mitsuru Kusumoto, Keisuke Yahata, Masahiro Sakai \\
Preferred Networks, Inc. \\
\{mkusumoto,yahata,sakai\}@preferred.jp}
\nocopyright  
\maketitle

\newcommand{\todo}[1]{{\color{MidnightBlue}{(TODO: #1)}}}

\newcommand\knote[1]{\textcolor{blue}{#1}}
\newcommand\qnote[1]{\textcolor{red}{#1}}
\newcommand\snote[1]{\textcolor{purple}{#1}}

\begin{abstract}
The problem-solving in automated theorem proving (ATP) can be interpreted as a search problem where the prover constructs a \emph{proof tree} step by step.
In this paper, we propose a deep reinforcement learning algorithm for \textit{proof search} in intuitionistic propositional logic.
The most significant challenge in the application of deep learning to the ATP is the absence of large, public theorem database. We, however, overcame this issue by applying a novel data augmentation procedure at each iteration of the reinforcement learning.
We also improve the efficiency of the algorithm by representing the syntactic structure of formulas by a novel compact graph representation.
Using the large volume of augmented data, we train highly accurate graph neural networks that approximate the value function for the set of the syntactic structures of formulas.
Our method is also cost-efficient in terms of computational time.
We will show that our prover outperforms Coq's \texttt{tauto} tactic, a prover based on human-engineered heuristics.
Within the specified time limit, our prover solved 84\% of the theorems in a benchmark library, while \texttt{tauto} was able to solve only 52\%.
\end{abstract}
\section{Introduction}

Deep learning has great success records for the planning problems over a discrete state space, with the celebrated AlphaGo being a prominent example.
AlphaGo won landslide victories against top-level human players in the game of Go \cite{silver2016mastering,silver2017mastering}, and its techniques were successfully extended to chess and shogi \cite{silver2017mastering2}.

One of the essential features of their work is the use of neural network-based evaluation functions that predict \textit{how close a given state is to the winning state}.
Neural networks are trained either by supervised learning based on a \textit{game record database} of human players, or by reinforcement learning through self-play.
At each step of the algorithm, trained neural networks are used to determine which states shall be explored by the \textit{search algorithm}.

Automated theorem proving (ATP) is a field that aims to prove formal mathematical theorems by the computer, and it has various applications such as software verification.
ATP can be seen as a symbolic reasoning-based planning problem in a discrete state space.
Despite recent improvement in general ATP systems and the development of specialized provers (such as SMT solvers) for numerous problems with practical social and scientific utility,
there still remain many problems that the current state-of-the-art solvers cannot prove.

In this paper, we would use neural networks to solve the problems of ATP,
with a hope that high generalization ability of neural networks can enable the automated proving of unseen complex problems in formal mathematics.

Indeed, our research is not the first to use neural networks in the realm of ATP.
However, only a few studies quantitatively compare their performance against conventional ATP systems based on human-engineered heuristics.
We must say there is not enough evidence to conclude that neural network is useful for solving ATP problems.

One evident difficulty in solving ATP problems using neural networks is the shortage of training data.
Most existing works rely on supervised learning that uses human annotated data (i.e., pairs of mathematical statements and their proofs).
Since training deep neural networks usually requires a large volume of data,
the use of only human annotated data might not be sufficient for the extraction of hidden features in complex mathematical statements.

Another difficulty is the severe trade-off between computation time and the quality of evaluation functions.
With recent advances in computational hardware, naive brute force algorithms can search thousands of states within a short time frame. For the neural network-based evaluation function to be useful, it must be able to yield an estimate with much higher accuracy than the brute force algorithms within the same time frame. 

In this paper, we propose a deep reinforcement learning algorithm for automated theorem proving in intuitionistic propositional logic (IPL).
The statement proving process of \textit{constructing a proof tree using inference rules of a sequent calculus} can be formalized as a search problem in a discrete state space.
We introduce neural networks to represent evaluation functions that determine which states to explore during the search.

Our algorithm assumes that we have access to a theorem library (problem set) containing mathematical statements for the training of evaluation functions.
In this paper, we train evaluation functions using a library containing up to only several thousand theorems since the size of all publicly available libraries today is of this scale.
In this paper, we deal with formulas that are several hundred in length. The traditional algorithms based on human heuristics we implemented in this study could not consistently handle the length of this range.

Our algorithm iteratively improves evaluation functions through approximate policy iteration (API) \cite{lagoudakis2003least}.
At each iteration, we apply data augmentation procedure to generate millions of data from a small theorem library, which was empirically sufficient for the training of neural network with strong generalization ability. 

The architecture of neural network to use is a critical factor for highly accurate prediction.
We use graph neural networks (GNNs) to represent evaluation functions.
One advantage of GNNs is that they can capture syntactic structures and the set of features that is independent of the variable nomenclatures.
Further, we propose a novel graph input format for GNNs that improves the prediction accuracy.

In experiments, we compare our prover against two provers that are based on sequent calculus:
Coq's automation tactic \texttt{tauto} and 
an untrained prover equipped with minimal heuristics.
We show that our algorithm outperforms both baseline provers.

The contributions of this paper are threefold:
\begin{itemize}
    \item 
        We propose an API-based reinforcement learning algorithm for ATP.
        We succeeded in training neural networks with high evaluation accuracy using a small theorem library by applying data augmentation.
    \item
        We show that, while slightly heavier in terms of computational cost, the GNNs-based evaluation functions are more accurate than those based on other network architectures.
    \item We show that our prover outperforms the baseline provers in terms of the number of theorems proven within the same time limit.
\end{itemize}

Some readers may consider propositional logic to be less interesting
because mathematical statements in IPL are much more restrictive than in predicate logic.
However, despite its restricted expressiveness,
the search space of IPL can exponentially grow with the length of a problem instance.
Also, the choice of the inference rules to use during the search algorithm---which corresponds to the choice of actions in reinforcement learning---plays a critical role in the development of an efficient ATP for IPL.
Thus, while simple and concise, ATP in IPL has enough complexity to potentially serve as a field of benchmark study for the future ATP research.
\section{Preliminaries}

\subsection{Intuitionistic Propositional Logic}

\textit{Propositional logic} is a logical system consisting of propositional variables and logical symbols ($\bot$ (false), $\land$ (and), $\lor$ (or), $\rightarrow$ (implication), and $\neg$ (negation)).
We denote propositional variables by uppercase letters $P$ and $Q$.
We assume that the reader is familiar with the syntax and the semantics of these symbols.
Any expression consisting of propositional variables and logical symbols is a \textit{formula}.
For example, ``$(P \lor Q) \rightarrow \neg P$'' is a formula.
We denote formulas by uppercase letters, $A$, $B$, and $G$.
Unless otherwise stated, we will treat $\neg A$ as $A \rightarrow \bot$ for any formula $A$.
The number of propositional variables and symbols in any given formula (where multiple appearances of the same variable or symbol are all counted) is referred to as the \emph{length} of the formula.
\emph{Proof} of a formula consists of applications of inference rules and axioms.
In this paper, we say that a formula is \emph{provable} if there exists a proof to the formula.
On the other hand, we say that a formula is \emph{solvable} by a prover if the prover can find a proof to the formula.
In the next section, we will more precisely explain what it means to \textit{be able to find a proof to the formula}.

\textit{Intuitionistic logic} is a logical system that is significantly different from the ``normal'' logic system to which we are accustomed in a standard mathematical argument.
``Normal'' logic is formally called \textit{classical logic}.
Classical logic assigns a boolean value to any arbitrary formula;  Intuitionistic logic does not.
A formula holds in intuitionistic logic if the formula is provable.
A formula that holds in classical logic may not hold in intuitionistic logic.
In particular, the law of excluded middle ($A \lor \neg A$) and double negation elimination ($\neg \neg A \rightarrow A$) do not necessarily hold in intuitionistic logic.
In this paper, we deal with \textit{intuitionistic propositional logic} (IPL).
For further details on IPL, please refer to a textbook on Logic \cite{buss1998handbook}.
\begin{figure*}[tbh]
\centering

\begin{tabular}{cccc}
\begin{bprooftree}
\AxiomC{}
\RightLabel{Init}
\UnaryInfC{$A, \Gamma \fCenter \ \Rightarrow\  A$}
\end{bprooftree}
&
\begin{bprooftree}
\AxiomC{}
\RightLabel{$\bot$-Left}
\UnaryInfC{$\bot, \Gamma \fCenter \ \Rightarrow\  G$}
\end{bprooftree}
&
\begin{bprooftree}
\AxiomC{$A,B,\Gamma \fCenter \ \Rightarrow\ G$}
\RightLabel{$\land$-Left}
\UnaryInfC{$A\land B, \Gamma \fCenter \ \Rightarrow\  G$}
\end{bprooftree}
&
\begin{bprooftree}
\AxiomC{$\Gamma \fCenter \ \Rightarrow\ A$}
\AxiomC{$\Gamma \fCenter \ \Rightarrow\ B$}
\RightLabel{$\land$-Right}
\BinaryInfC{$\Gamma \fCenter \ \Rightarrow\ A\land B$}
\end{bprooftree}
\end{tabular}

\begin{tabular}{ccc}
\begin{bprooftree}
\AxiomC{$A, \Gamma \fCenter \ \Rightarrow\  G$}
\AxiomC{$B, \Gamma \fCenter \ \Rightarrow\  G$}
\RightLabel{$\lor$-Left}
\BinaryInfC{$A\lor B, \Gamma \fCenter \ \Rightarrow\  G$}
\end{bprooftree}
&
\begin{bprooftree}
\AxiomC{$\Gamma \fCenter \ \Rightarrow\  A$}
\RightLabel{$\lor$-Right1}
\UnaryInfC{$\Gamma \fCenter \ \Rightarrow\  A\lor B$}
\end{bprooftree}
&
\begin{bprooftree}
\AxiomC{$\Gamma \fCenter \ \Rightarrow\  B$}
\RightLabel{$\lor$-Right2}
\UnaryInfC{$\Gamma \fCenter \ \Rightarrow\  A\lor B$}
\end{bprooftree}
\end{tabular}

\begin{tabular}{cc}
\begin{bprooftree}
\AxiomC{$A\rightarrow B, \Gamma \fCenter \ \Rightarrow\  A$}
\AxiomC{$B, \Gamma \fCenter \ \Rightarrow\  G$}
\RightLabel{$\rightarrow$-Left}
\BinaryInfC{$A\rightarrow B, \Gamma \fCenter \ \Rightarrow\  G$}
\color{black}
\end{bprooftree}
&
\begin{bprooftree}
\AxiomC{$A, \Gamma \fCenter \ \Rightarrow\ B$}
\RightLabel{$\rightarrow$-Right}
\UnaryInfC{$\Gamma \fCenter \ \Rightarrow\ A\rightarrow B$}
\end{bprooftree}
\end{tabular}
\caption{
    Inference rules in sequent calculus LJ from \cite{dyckhoff1992contraction}.
    Note that these rules are cut-free.
    \label{fig:LJ}}
\end{figure*}

\subsection{Sequent Calculus LJT}
 
In this paper, we use sequent calculus as inference rules.
We will first explain LJ, a standard sequent calculus formulation.
We will then explain LJT \cite{dyckhoff1992contraction}, a variant of LJ that is more suitable for automated reasoning.

Inference rules in sequent calculus LJ are shown in Figure~\ref{fig:LJ}.
In the figure, $A, B$, and $G$ each represent a single formula, and $\Gamma$ represents a multiset of formulas.
$\Gamma$ may be empty or may contain multiple formulas.
An expression of the form ``$A_1, \ldots, A_k \Rightarrow G$'' is called a \textit{sequent}, 
with the left hand side ($A_1, \ldots, A_k$) standing for \textit{antecedents}, and the right hand side ($G$) standing for \textit{consequent}.
The sequent is semantically equivalent to ``$A_1 \land \ldots \land A_k \rightarrow G$''.
Proving a formula $A$ is equivalent to proving the sequent ``$\Rightarrow A$''.

In each inference rule in Figure~\ref{fig:LJ},
an upper part of a horizontal line is a \textit{premise(s)} to derive the \textit{conclusion} in the lower part.
There may be no premises, or there may be multiple premises.
We must prove all the premise sequents to derive the conclusion.
An inference rule with no premises works as an axiom.
Note that there are no inference rules for the $\neg$ symbol.
This is because we can write $\neg A$ as $A \rightarrow \bot$, so that the inference rules for $\rightarrow$ suffices.

We can visualize the process of a proof as a \textit{proof tree}.
For example, we can prove the sequent ``$\Rightarrow P \land Q \rightarrow Q \land P$'' by the proof tree presented below.

\begin{prooftree}
\AxiomC{}
\RightLabel{Init}
\UnaryInfC{$P, Q \ \Rightarrow\ Q$}
\AxiomC{}
\RightLabel{Init}
\UnaryInfC{$P, Q \ \Rightarrow\ P$}
\RightLabel{$\land$-Right}
\BinaryInfC{$P, Q \ \Rightarrow\ Q \land P$}
\RightLabel{$\land$-Left}
\UnaryInfC{$P \land Q \ \Rightarrow\ Q \land P$}
\RightLabel{$\rightarrow$-Right}
\UnaryInfC{$\Rightarrow\ P \land Q \rightarrow Q \land P$}
\end{prooftree}

In any proof tree, the goal sequent must be placed at the bottom.
The top leaves in each branch must all be applications of axioms (i.e., Init or $\bot$-Left).

With sequent calculus LJ, we can \textit{prove} a goal sequent by constructing a proof tree from bottom to top.
The construction ends when all the top leaves in a proof tree are applications of axioms.
We can formalize this process as a search problem.
We will give the formal definition in the next section.

An inference rule $\rightarrow$-Left in LJ may trigger undesirable behavior for automated reasoning
since the application of $\rightarrow$-Left may elongate a sequent without a bound.
Put in more practical terms, we cannot guarantee the termination of proof search in LJ.
Sequent calculus LJT is a modified version of LJ,
where $\rightarrow$-Left is replaced with another inference rule so we can guarantee that the length of any sequent will never increase after the application of any inference rules \cite{dyckhoff1992contraction}.
Please refer to the original paper for the details of LJT.
In this paper, we use LJT for theorem proving.

\subsubsection{Choice of inference rules during proof search}
The choice of inference rules allowed in the search algorithm is crucial for the efficient construction of proof tree.
For example, suppose that we have a sequent ``$\Rightarrow A \lor B$'' where only either one of $A$ and $B$ is provable.
With a relatively straightforward argument, one can deduce that $\lor$-Right1 and $\lor$-Right2 are the only two inference rules that can be applied to this sequent.
One of them will lead to provable sequent and the other will lead to unprovable sequent.
If we make a wrong decision, we will consume unnecessary time in vain for searching a proof of the unprovable sequent. 

Indeed, depending on the situation, the choice may not be always so crucial.
For example, if the original sequent is ``$A\land B \Rightarrow A \land B$'', the applicable inference rules are  $\land$-Left,  $\land$-Right, and Init, and it turns out that none of them changes the logical validity of the sequent, although they may increase the necessary steps to end the proof.

\subsubsection{Theorem proving in IPL}
Theorem proving in IPL is a well-studied subject in Logic \cite{raths2007iltp}.
Determining whether a given formula is provable in IPL is PSPACE-complete and is therefore NP-hard, which implies that, in the worst case, the search space grows exponentially with the length of formula (unless, of course, P=NP.)

For a benchmark to compare against our prover, we used \texttt{tauto}, a proof automation tactic based on a variant of LJT that is featured in Coq proof assistant \cite{coq}, which is a widely used interactive theorem prover.

There exist calculi other than sequent calculus LJT that are more efficient for the theorem proving in IPL.
For example, fCube \cite{ferrari2010fcube}, an automated theorem prover in IPL based on tableau calculus,
can prove most of the theorems we deal with in this paper within milliseconds.
In this paper, we use sequent calculus LJT since sequent calculus is easier to understand than tableau calculus.
In order to avoid the debate regarding the efficacy of the calculus, we would restrict our study to the set of provers based on LJT.
\subsection{Tasks}

We assume that we are given a pair of theorem libraries (problem sets of provable formulas in IPL), $D_\mathrm{train}$ and $D_\mathrm{exam}$.
We use $D_\mathrm{train}$ for the training of neural networks during the reinforcement learning.
We use $D_\mathrm{exam}$ to evaluate the performance of the trained prover.
We assume that $D_\mathrm{exam}$ is not available during the reinforcement learning.
In our experiment, we deal with formulas that are several hundred in length.

To evaluate the efficacy of our method, we conducted two tasks: \textit{prediction task} and \textit{proving task}.
In the prediction task, we compared our GNNs against other neural network architectures in terms of their abilities to approximate the value function.
We performed supervised learning for the training of each architecture and reported the test error the inference time.

In the proving task, we compare the performance of our prover against baseline provers. 
We also demonstrate that our prover improves with each iteration of the API algorithm.
\section{Method}

In this section, we explain how our method develops a prover through reinforcement learning.
For the detailed description of terminologies and methods in reinforcement learning, please consult a textbook such as \cite{sutton1998introduction}.

\begin{figure}[t!]
  \centering
  \includegraphics[width=8cm]{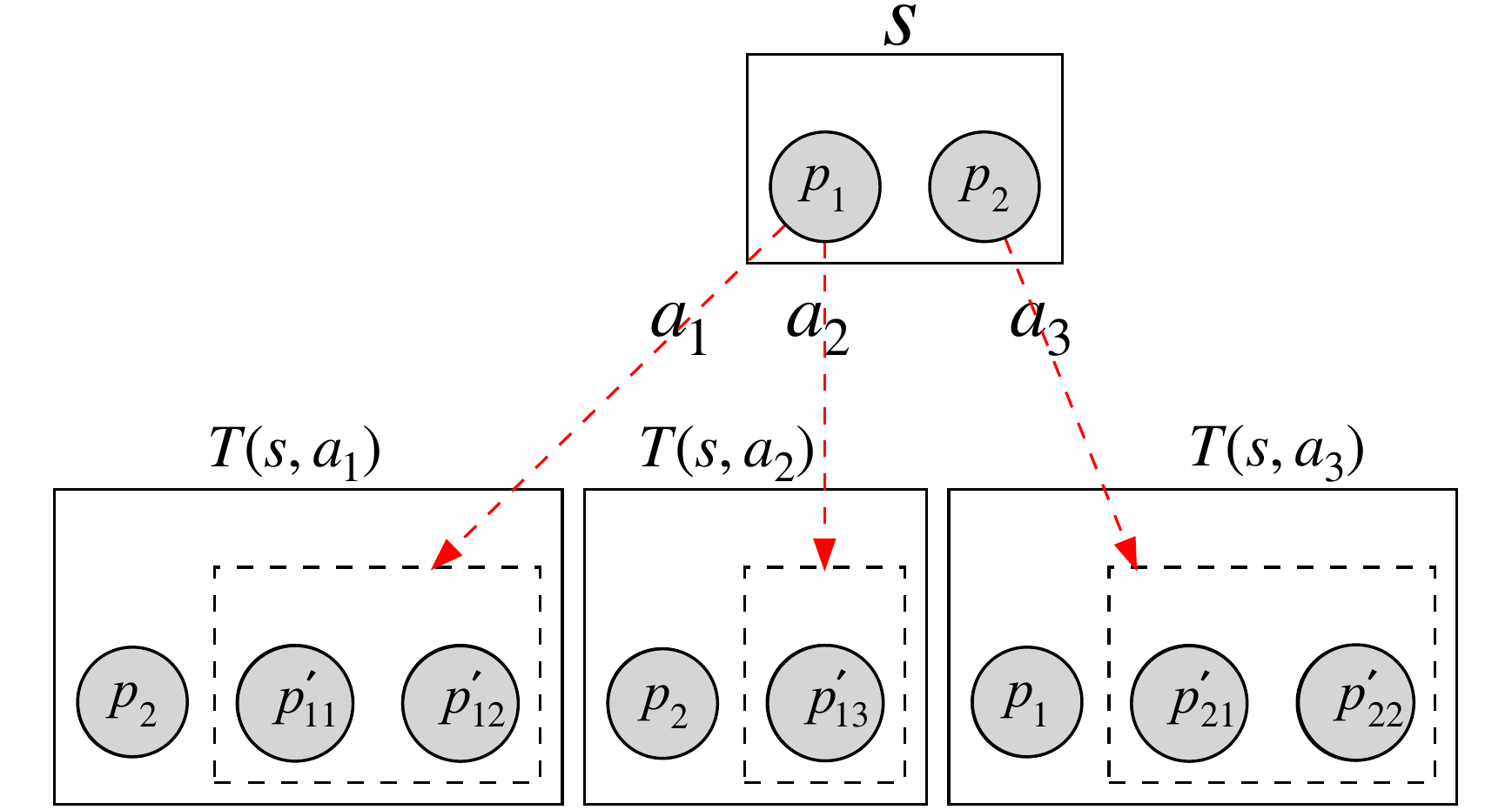}
  \caption{An overview of the Markov decision process.
  \label{fig:MDP}}
\end{figure}

\subsection{Markov Decision Process Formulation}

The objective of \textit{proof search} is to prove a formula using inference rules in LJT.
In other words, if ``$A$'' is the formula to be proven, our objective amounts to the construction of a proof tree that has ``$\Rightarrow A$'' at the bottom.
We formalize this problem as a Markov decision process (MDP), a tuple $(\mathcal{S}, \mathcal{A}, T, r, \gamma)$ consisting of the following five elements:

\begin{itemize}
    \item $\mathcal{S}$ is the set of \textit{states}.
        At any stage in the construction of proof tree, we only need to keep track of the set of the sequents that remains to be proven. Our state space $\mathcal{S}$ is, therefore, the powerset of all sequents.
        The empty state $\emptyset \in \mathcal{S}$ corresponds to the end of the proof because it literally represents the state at which there are no more sequents to be proven.
        Throughout this paper, we denote a state in $\mathcal{S}$ by $s$ and a single sequent by $p$.
    \item $\mathcal{A}_s$ is the set of \textit{actions} permitted at the state $s$, and $\mathcal{A}$ is the collection of $\mathcal{A}_s$.
        \textit{An action} in our MDP is an application of an inference rule to a sequent in $s$.
        $\mathcal{A}_s$ varies across different states because the set of the inference rules applicable to a sequent depends on the sequent.
        Each $a \in \mathcal{A}_s$ is a pair $(p, r)$ consisting of a sequent $p \in s$ and an inference rule $r$ permitted for $p$. Upon the application of the action $a = (p, r) $, the inference rule $r$ is applied to the sequent $p$. That is, $s$ would be updated to the next state $s'$ of the form $s' = (s \setminus \{p\}) \cup \{p'_1, \ldots, p'_k\}$, where $p'_1, \ldots, p'_k$ are the sequents in premises of the inference rule $r$.
        
        Note that $\mathcal{A}_s$ may be empty.
        This happens either at the successful end of the proof or at the logical stalemate.
    \item $T$ is the \textit{transition function} that maps a state-action pair to the next state.
        For a state $s$ and an action $a$, we define $T(s, a) := s'$,
        where $s'$ is the state defined by $s$ and $a$ as above.
    \item $r:\mathcal{S} \to \mathbb{R}$ is the \textit{reward function}.
        When a prover reaches state $s$, the prover will receive the reward $r(s)$.
        We define $r(\emptyset)=1$ and $r(s)=0$ $(s \not= \emptyset)$.
        This is to say that the prover earns non-zero reward only upon the successful completion of the proof.
    \item $\gamma \in (0, 1)$ is the \textit{discount-rate}.
        This value is used to compute the return, which we will explain momentarily.
\end{itemize}
The MDP for a proof search is illustrated in Figure~\ref{fig:MDP}.
Under this formulation of the MDP, our objective can be written as a problem of finding the optimal \textit{policy} function $\pi^*$ that maps a given state $s \in \mathcal{S}$ to the action $\pi^*(s) \in \mathcal{A}_s$ that is optimal at $s$.
In general, we assume that a policy is a deterministic function.

Given a policy $\pi$ and an initial state $s$,
an \textit{episode} is a sequence of states and actions $(s_0, a_0, s_1, a_1, s_2, \ldots)$, where $s_0 = s$, $a_i = \pi(s_i)$, and $s_{i+1} = T(s_i, a_i)$.
An episode ends when no more possible actions are available, i.e., $A_{s_n} = \emptyset$.
The \textit{return} $R(s)$ of an episode is defined as follows: $R(s) = \sum_{i \ge 0} \gamma^i r(s_i)$.
Note that the return $R(s)$ depends on the policy $\pi$.
Our objective here is to find the policy $\pi$ that maximizes the expected return when the initial state $s$ is randomly sampled from the distribution governing the training dataset (e.g., theorem library);
\[
\maximize_{\pi} \E_s [R(s)].
\]

In our formulation of the MDP,
we can simplify the calculation of the return $R(s)$ since the \textit{prover(agent) earns} a positive reward only when the policy finishes the proof.
If the episode starting from an initial state $s$ ends successfully in $n$ steps,
i.e., $s_n = \emptyset$,
then $R(s) = \gamma^n$.
If the episode ends unsuccessfully, then $R(s) = 0$.
In both cases, we can write $R(s) = \gamma^n$ if we regard $n=\infty$ for the unsuccessful termination of the proof. 
Also, consider a state $s \in \mathcal{S}$ consisting of sequents $p_1, \ldots, p_k$.
Since the proof of each sequent $p_j$ is independent,
the required number of steps to prove $s$ is the sum of the number of steps $n_j$ required to prove each $p_j$.
This implies that $R(s) = \gamma^{n_1+\cdots +n_k} =   R(\{p_1\}) \times \cdots \times R(\{p_k\})$.

\begin{figure}[t!]
  \centering
  \includegraphics[width=7cm]{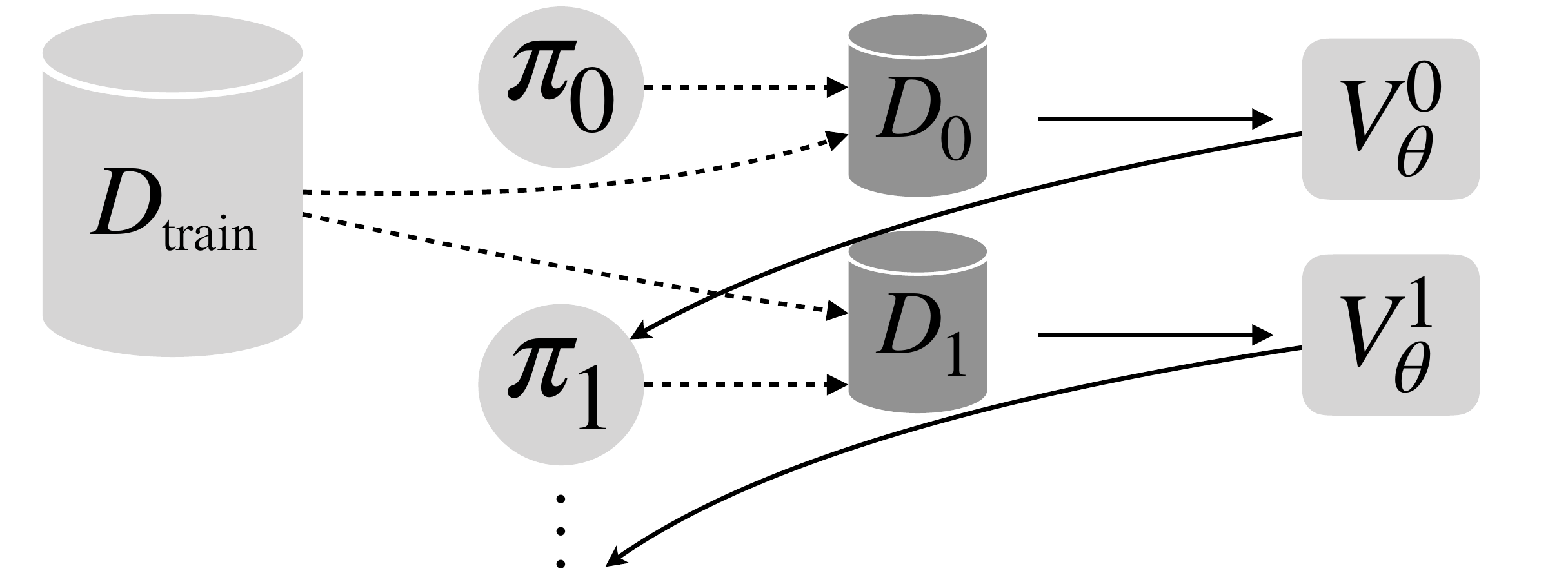}
  \caption{A pipeline of approximate policy iteration.
  \label{fig:policy-iteration}}
\end{figure}
\begin{figure*}[htb]
  \centering
  \includegraphics[width=16cm]{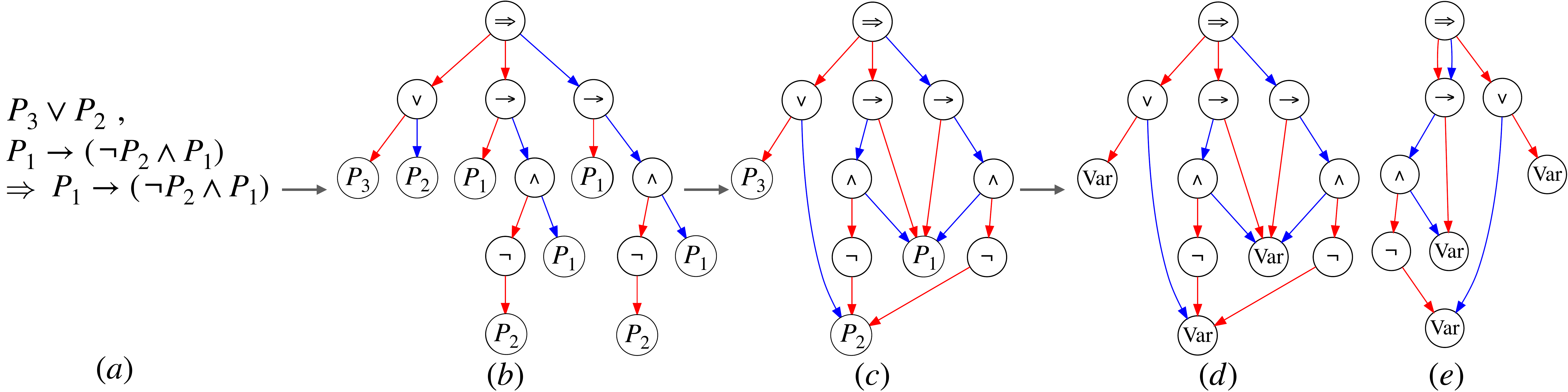}
  \caption{Conversion from a sequent to a directed graph.
  (a) The input sequent.
  (b) The abstract syntax tree of the sequent with edge labels.  The label ``edge-left'' is colored by red, the label ``edge-right'' is colored by blue.
  (c) Merging of the same propositional variable vertices.
  (d) Replacement of the names of propositional variable by a common label ``Var''. The steps  (a) $\sim$ (d) constitutes the variable-merging (VM) process.
  (e) The result of term-merging (TM). Note that two appearances of ``$P_1 \rightarrow (\neg P_2 \land P_1)$'' are merged in this graph.
  \label{fig:graph-input}}
\end{figure*}

\subsection{Approximate Policy Iteration}

In the search for the optimal policy $\pi^*$, we perform approximate policy iteration (API),
an iterative algorithm that incrementally updates policies starting from an arbitrary baseline policy $\pi_0$.

In the standard literature of reinforcement learning, a value function $V^\pi : \mathcal{S} \to \mathbb{R}$ is defined as the expectation of the return for policy $\pi$, which can be stochastic.
In our setting, both the transition functions and policies are deterministic, so there is no stochasticity in the above definition.
That is, $V^\pi(s) :=  \sum_{i \ge 0} \gamma^i r(s_i) = R(s)$.

In (non-approximate) policy iteration, at every $i$th episode we are required to compute the value function $V^{\pi_i}$ for policy $\pi_i$ by evaluating the return for all $s \in \mathcal{S}$.
After obtaining the value function, we update the policy from $\pi_i$ to $\pi_{i+1}$ through the following rule.
\begin{equation}
\label{eq:def-pi}
\pi_{i+1}(s) := \argmax_{a \in \mathcal{A}_s} V^{\pi_i}(s') \quad (s' := T(s, a)).
\end{equation}
Starting from a baseline policy $\pi_0$, we compute the value functions and policies alternately; That is, we will be computing $V^{\pi_0}$, $\pi_1$, $V^{\pi_1}$, $\pi_2$, $V^{\pi_2}$, and so on, in this order.

In our problem setting, however, the exact evaluation of the value function $V^{\pi_i}$ is impossible, because the cardinality of $\mathcal{S}$ is infinity. We therefore approximate the value function using a neural network parametrized by $\theta$, which we denote by $V^{\pi_i}_\theta$. 
This method is called approximate policy iteration (API).
In each iteration, we compute the approximated value function $V^{\pi_i}_\theta$ through supervised learning.
We compute the output of the policy $\pi_{i+1}(s)$ using Equation~\ref{eq:def-pi}.
In API, there is no guarantee that we can improve policies due to approximation errors.
As in many other studies based on API, however, we expect on a heuristic basis that we can improve the policies in each iteration.

For any given state $s = \{p_1, \ldots, p_k\} \in \mathcal{S}$,
we can write $V^\pi(s) = V^\pi(\{p_1\}) \times \cdots \times V^\pi(\{p_k\})$.
To simplify the problem, we assume that the following analogous identity holds for the approximated value function:
\[
V^{\pi}_\theta(s) := V^\pi_\theta(\{p_1\}) \times \cdots \times V^\pi_\theta(\{p_k\}).
\]
For ease in analysis, we will therefore be writing all approximated value function in terms of $V^\pi_\theta (p)$ only.

To obtain $V^{\pi_i}_\theta$,
we first generate a dataset $D_i$ from $D_\mathrm{train} $ that consists of pairs $(p, R(p))$, where $p$ is a sequent and $R(p)$ is the return.

We then use this dataset to approximate $V^{\pi_i}$ by the neural network function $V^{\pi_i}_\theta$ in a supervised manner. More precisely,
we seek for the parameter $\theta$ that minimizes the mean squared error between the predicted value and the actual return: 
\begin{equation}
\label{eq:loss}
    \frac{1}{|D_i|} \sum_{(p, R(p)) \in D_i} \Bigl( V^{\pi_i}_\theta(p) - R(p) \Bigr)^2.
\end{equation}
\subsection{Data Augmentation Procedure}

As mentioned in the introduction, the existing theorem libraries are too small to train neural networks with high generalization ability. We therefore propose a novel data augmentation method for the theorem library. We will provide a brief sketch of the procedure in this section and leave the technical details to the Supplemental Material.

Application of inference rule to unsolvable formula will most likely result in unsolvable formulas. We shall, therefore, generate augmented datasets from the set of solvable sequents. Our data augmentation procedure begins by iterating over the set of sequents in the library to determine which sequents are solvable by the current policy $\pi$. That is, for each sequent $p$, we compute an episode $(s_0, a_0, s_1, a_1, \ldots)$ with $s_0=\{p\}$ and check whether the final state $s_n$ is equal to $\emptyset$. 
The set of sequents considered unsolvable in the first step will not be included in the dataset with which we will update the current policy. 
We may say that this procedure is a variant of curriculum learning because our algorithm starts from learning easy problems and moves on to learn more and more challenging problems at every step.

To generate the augmented dataset from a sequent $p$,
we perform the breadth-first search (BFS) starting from $p$.
For each intermediate sequent $p'$ encountered during the BFS, we check for all sequents reachable from $p'$ using the set of inference rules permitted for $p'$. 
Note that we perform the BFS over the sequents, not over the state space $\mathcal{S}$ defined in the MDP.
We chose to use the BFS for this procedure since we wanted to generate sequents that are similar to the original sequent $p$.
In contrary, other search algorithms such as depth-first search (DFS) may generate sequents that are far from the original sequent $p$.
Note that some augmented sequents may be unprovable and some provable; this shall promote the diversity of the dataset.

During the BFS,
we sometimes encounter a situation in which we generate many sequents that can be solved with an application of only one inference rule (i.e., Init or $\bot$-Left).
We call such a sequent \textit{one-step-provable}.
While some amount of one-step-provable sequents is useful in terms of the diversity of the dataset,
the excessive presence of trivial data is detrimental for the training.
We therefore threshold the number of one-step-provable sequent in the dataset at $N_\mathrm{=1}$.
We terminate the BFS if the number of the generated data reaches the threshold $N_\mathrm{\ge 2}$.
\subsection{Graph Neural Networks}

In this paper, we use \textit{graph neural networks} (GNNs) to construct an approximate value function $V^\pi_\theta$.
Since logical formulas follow syntax rules, they can be represented by graphs (abstract syntax tree).
GNNs can directly work on this graph structure to extract hidden features.

Another advantage of GNNs is that it can extract hidden features that are invariant under variable renaming.
Logical formulas have a property that renaming of the propositional variables does not change their semantics. 
For example, two sequents ``$\Rightarrow P_1 \lor P_2 \rightarrow P_1$'' and ``$\Rightarrow Q_8 \lor Q_4 \rightarrow Q_8$'' contain variables with different names, but they have identical provability.
We can design GNNs that always output the same value for such an equivalent pair of formulas.

In this paper, we use gated graph neural networks~\cite{DBLP:journals/corr/LiTBZ15} as GNNs.
To evaluate a \textit{sequent} with gated GNNs, we first convert the sequent to a directed graph with vertex and edge labels. Here, we follow a similar method to the one proposed by \cite{wang2017premise}.
This process is summarized in Figure~\ref{fig:graph-input}.
We begin by converting the input sequent~(a) to its abstract syntax tree~(b), where the leaf vertices are propositional variables with variable names (e.g. $P_1, P_2$ ).
The edges are labeled by ``left''(red) or ``right''(blue), depending on the relative location of the terms flanking the parent operator (e.g. $\land$). 
We then merge the leaf vertices with the same name. This will result in a graph like~(c). Finally, we remove the variable names from the leaves and instead assign the common label ``Var'' to all variables.
We refer to the resulting graph~(d) as a graph represented in \textit{variable-merging} (VM) format.

After the conversion of the formulas to the graphs,  we compute the hidden features on each vertex of the graph iteratively.
First, we assign a feature vector $h^0_v$ to each vertex $v$, given by the one-hot vector representing the $v$'s vertex label.
We compute the new embeddings $h^t_v$ from the previous embeddings of vertices adjacent to $v$ using the following rule (we denote a set of directed edges in the graph by $E$).
\[
    h^t_v = F(h^{t-1}_v,
              \{ h^{t-1}_{w^-} \}_{w^- : (v, w^-) \in E},
              \{ h^{t-1}_{w^+} \}_{w^+ : (w^+, v) \in E}).
\]
Here, $F$ is a non-linear function containing the parameters that are subject to optimization.
After $T$ steps, we obtain the embeddings $\{h^T_v\}_v$.
We define the output of the GNNs as $G(\sum_v h^T_v)$,
where $G$ is a non-linear function with trainable parameters that maps the input to real numbers.
\subsection{Improving the Graph Input Format}

We can further compress the graph in the VM format by merging the collection of equivalent \textit{terms}.
This will result in a graph of format~(e).
We call this format \textit{term-merging format} (TM).
The graph~(e) in Figure~\ref{fig:graph-input} is an example of a graph in the term-merging format.

Term-merging makes the semantic relation in the formula clearer than variable-merging.
For example, when the sequent is immediately provable by the inference rule \textit{Init},
one of the left-edges and the right-edge from the root vertex ($\Rightarrow$) points to the same vertex, as in Figure~\ref{fig:graph-input}~(e).
This property makes it easier for the GNNs to determine whether the sequent is one-step-provable.
In the experimental section, we will show that the TM format is not only more compact than the VM format, but also that the data in this format yields predictor with better accuracy.
\subsection{Search Algorithm with Backtracking}

The action set $\mathcal{A}$ defined for our MDP formulation does not allow backtracking.  As such, there is always a danger that the prover gets stuck at $\mathcal{A}_s=\emptyset$.  
To improve the performance of our prover on $\mathcal{D}_{\textrm{exam}}$,
we introduce another search algorithm \textit{greedy DFS},
which performs the depth-first search with the action set $\mathcal{A}$.
During the DFS, we make the prover choose the actions that direct the prover to the states with higher value function scores.

Further, we slightly modify the action set $\mathcal{A}$ to reduce the number of actions.
For any state $s=\{p_1, \ldots, p_k\}$, we re-define $\mathcal{A}_s$ as the set of all actions allowed for the sequent $p_{j^*}$,
the sequent with the lowest value assigned by the evaluation function among all sequents in $s$. 
Recall that, after $p_{j^*}$ is solved, the new state $s'$ is devoid of $p_{j^*}$.  Therefore, if $s$ is provable, a prover equipped with this \textit{new} definition of action can still solve $s$.  

Also, to reduce the computation time, we use batch-computation when we calculate Equation~\eqref{eq:def-pi}.
Further, we cache the inference result of the value functions so that we can omit the evaluations for redundant input sequents.
\section{Experiments}

In this section, we will briefly describe the setup of our experiments for \textit{the prediction task} and \textit{the proving task} (see Preliminaries section). For more technical details of the experiments, please see the Supplemental Material.

Throughout, our baseline policy is a \textit{naive greedy} policy that chooses an action that minimizes the total length of the next state:
\[
    \pi_0(s) := \argmin_{a \in \mathcal{A}_s} \sum_{p' \in s'} \mathrm{length}(p'), 
\]
where a length of a sequent is defined as the total length of formulas in the sequent.
\subsection{Prediction Task}

\begin{table}[t!]
  \centering
  \begin{tabular}{|l|r|r|} \hline
    Architecture & Test error ($D_0$) & Inference Time \\ \hline
    BoW      & 0.142 & 0.3ms \\
    LSTM     & 0.088 & 2.4ms \\
    GRU      & 0.083 & 2.4ms \\ \hline
    GNN (VM) & 0.043 & 3.8ms \\
    GNN (TM) & \textbf{0.036} & 3.8ms \\ \hline
  \end{tabular}
  \caption{Performances of different neural network architectures. We are reporting the mean-squared error in Equation~\eqref{eq:loss} as the test error. \label{table:models}}
\end{table}
For the prediction task, we used three network architectures to predict the \textit{return} on the dataset $D_0$ generated by $\pi_0$, and compared their accuracies: bag-of-words (BoW), recurrent neural network (RNN), and the gated GNN. Within the category of RNNs, we 
experimented with two variants: GRU and LSTM.
We describe the details of these architectures in Supplemental Material.

The augmented dataset $D_0$ contained 851393 sequents. We split $D_0$ into the three datasets---train, validation, and test set---with a respective ratio of 4:1:1.
In order to guarantee that all three datasets are perfectly independent from one another, we made sure that all sequents generated from the same original sequent are included in only one of the three sets.

For the comparison of the performance of the architectures, we used the mean squared error from Equation~\eqref{eq:loss}.

We present the result of the prediction task in Table~\ref{table:models}.
We see that GNNs are much more accurate than other baseline architectures, with the test error of around 0.04.  The test errors of the other architectures are greater than 0.08.
As for inference time, GNNs are only about 1.5 times slower than RNNs.
We observed that the large step size $T$ of GNNs was necessary for achieving high accuracy.
While GNNs with step size 1 were as poor as BoW, the increase in the step size continued to significantly improve the validation error until $T=6$.
As for the graph input format, we observed that term-merging (TM) is more effective than variable-merging (VM).
\subsection{Proving Task}

We assessed the performance of our prover in comparison to the baseline methods. We performed the reinforcement learning algorithm to obtain policies $\pi_1$, $\pi_2$, and so on, starting from the baseline policy $\pi_0$.
To demonstrate the effect of our data augmentation procedure, we also ran the approximate policy iteration without augmentation.
We present the result for this set of experiments in
Table~\ref{table:policy-iteration}.
The initial policy $\pi_0$ worked poorly, proving only 41\% of theorems in $D_\mathrm{train}$.
After two iterations of the approximate policy iteration, the policies almost converged, and $\pi_2$ could prove around 85\% of all theorems.
The number of training data $D_i$ increased with the iteration.
$D_3$ contained 1753080 data, which is roughly twice the size of $D_0$.
Note that our prover ran without backtracking during the API.

Without data augmentation, the policies quickly degraded.
After one iteration, the policy was able to prove only 24\% of the theorems.
This result shows the importance of data augmentation procedure.

\begin{table}[t!]
  \centering
  \begin{tabular}{|c||r|r|r|r|r|} \hline
    Iteration &$\pi_0$ & $\pi_1$ & $\pi_2$ & $\pi_3$ & $\pi_4$  \\ \hline
    With augmentation  & 41\% & 64\% & 85\% & 86\% & 86\% \\ \hline
    No augmentation    & 41\% & 24\% & 25\% & 24\% & 14\% \\ \hline
  \end{tabular}
  \caption{The percentages of theorems in $D_\mathrm{train}$ proven by the policies at different stages of API.
  \label{table:policy-iteration}}
\end{table}

Secondly, we compared our trained prover $\pi_4$ against two baseline provers with the dataset $D_\mathrm{exam}$: Coq's \texttt{tauto} tactic (version~8.7.1) and our baseline prover $\pi_0$ with greedy DFS.
Since the number of theorems that the provers can prove depends heavily on the time limit,
we conducted the comparative study using several different time constraints.

We present the result of the proving task in Table~\ref{table:coq}.
While the two baselines solved at most 52\% of all theorems in $D_\mathrm{exam}$ within the time limit of 10 seconds,
our trained prover succeeded in solving 60\% of all theorems within only 1 second.
Under all time constraints, our provers was able to solve more than 25\% of the theorems that the baseline provers could not.

We emphasize that our prover is playing this competition at a disadvantage because the two baseline provers are deploying heuristics that enable the decision making within a minuscule time frame. Our superiority in this comparative experiment, therefore, suggests that our prover is making efficient decisions during the proof search.     
For more details on the experimental results, please see the Supplemental Material.

\begin{table}[t!]
  \centering
  \begin{tabular}{|c||r|r|r|} \hline
    Prover & TL $\le 1$s & TL $\le 3$s & TL $\le 10$s  \\ \hline
    Coq's \texttt{tauto} & 35\% & 43\% & 52\%  \\
    Our baseline $\pi_0$ & 30\% & 35\% & 40\%  \\ \hline
    Our prover $\pi_4$   & \textbf{60\%} & \textbf{78\%} & \textbf{84\%}  \\ \hline
  \end{tabular}
  \caption{Comparison against baseline provers over exam library $D_\mathrm{exam}$. 
  \label{table:coq}}
\end{table}
\section{Related Work}

There have been several endeavors to combine existing provers and neural networks that are trained with human annotated dataset.
Loos~et~al. used neural networks for clause selection in \emph{E~Prover} and found proofs of 7\% of theorems in the Mizar library for which there have not been ATP generated proofs \cite{loos2017deep}.
Other studies in this category include algebraic rewriting problem in Coq interpreter \cite{huang2018gamepad}, higher-order logic in  Metamath library \cite{whalen2016holophrasm}, IPL \cite{sekiyama2018automated}, and SAT \cite{selsam2018learning}.
Another important application of neural network to ATP that is worthy of mention but is not directly related to \emph{proof search} algorithm is \emph{premise selection.} 
Premise selection is a preprocessing step for ATP that selects a set of lemmas from large theorem libraries that would become necessary in proof search. For applications of neural networks to premise selection, see, for example, \cite{irving2016deepmath,kaliszyk2017holstep,wang2017premise}.

Unfortunately, however, not many studies compare their methods to traditional provers on a quantitative basis. To the best of our knowledge, our research is the first of its kind in conducting an extensive study to show the effectiveness of the neural network-based algorithm in comparison to the traditional provers.

Moreover, the use of reinforcement learning in ATP is rare.
To name a few, Lederman~et~al. conducted REINFORCE algorithm for clause selection in a QBF solver using a graph neural network with step size one, and succeeded in
reducing the total steps relative to existing heuristic algorithms \cite{lederman2018learning}.
Also, Kaliszyk~et~al. trained a prover by reinforcement learning with XGBoost predictor based on manually engineered features, and showed that their prover can prove 40\% more theorems than their baseline prover in the Mizar library \cite{kaliszyk2018reinforcement}.

GNNs have been successfully applied to various fields such as biochemistry \cite{duvenaud2015convolutional} and program analysis \cite{allamanis2017learning}.
In the realm of logic, the use of GNNs has been limited to CNFs \cite{lederman2018learning} and the HolStep dataset \cite{wang2017premise}.
Our work provides new evidence for the efficacy of GNNs in the field of logic.

Evans~et~al. reports that recursive neural networks can accurately predict the entailment of formulas in classical propositional logic \cite{evans2018can}.
However, recursive networks are, known to be computationally inefficient, even when they are highly optimized \cite{bowman2016fast}.
\subsection*{Acknowledgement}
We thank Masanori Koyama, Shin-ichi Maeda, Toshiki Kataoka, Yasuhiro Fujita, and Yuu Sunahara in Preferred Networks, Inc. for technical suggestions.

\bibliography{references}
\bibliographystyle{aaai}

\newpage
\section{Supplemental Material}

In this section, we describe the details of our method and experimental results that we omitted from the main part of the paper.

\subsection{Detailed Data Augmentation Procedure}

We show the entire algorithm for our data augmentation procedure in Algorithm~\ref{alg:augmentation}.

\begin{algorithm}[htb]
\DontPrintSemicolon
\SetKwInOut{Input}{input}\SetKwInOut{Output}{output}

\Input{Policy $\pi$, library $D_\mathrm{train}$, and thresholds $N_{\ge 2}$ and $N_{=1}$.}
\Output{Generated dataset $D$.}
\BlankLine
\For{$p \in D_\mathrm{train}$}{
    Run policy $\pi$ to prove $p$.\;
    \If{$\pi$ could not prove $p$}{
        $D_p := \emptyset$.\;
        continue
    }
    \Comment{Run BFS from $p$}
    $Q := \{ p \}$. \Comment*{Queue of sequents}
    $D_{\ge 2} := \emptyset, D_{=1} := \emptyset$. \Comment*{Augmented data}
    \While{$Q$ is not empty and $|D_{\ge 2}| < N_{\ge 2}$}{
        $p' := Q$.pop(). \;
        \If{We already visited $p'$ during the BFS}{
            continue
        }
        Run policy $\pi$ to prove $p'$. \;
        $R(p')$ := Obtained return by running $\pi$. \;
        \If{$p'$ is one-step-provable}{
            \If{$|D_{=1}| < N_{=1}$}{
                $D_{=1} := D_{=1} \cup \{(p', R(p'))\}$.\;
            }
        }
        \Else{
            $D_{\ge 2} := D_{\ge 2} \cup \{(p', R(p'))\}$.\;
        }
        \For{$p''$ in possible next sequents from $p'$}{
            $Q$.push($p''$). \;
        }
    }
    $D_p := D_{\ge 2} \cup D_{=1}$.\;
}
\Return{$D:=\bigcup_{p \in D_\mathrm{train}} D_p$}

\caption{Data augmentation procedure. \label{alg:augmentation}}
\end{algorithm}

\begin{algorithm}[tb]
\DontPrintSemicolon
\SetKwInOut{Input}{input}\SetKwInOut{Output}{output}

\Input{Desired length $n$,
    number of variables $m$,
    probability vector $q=(q_\land, q_\lor, q_\rightarrow, q_\neg)$}
\Output{Formula $A$}
\BlankLine
\If{$n \le 2$}{
    $P$ := Choose a propositional variable from $m$ candidates uniformly at random.\;
    \Return{$P$}
}
$\mathit{op}$ := Choose an operator from $\{\land, \lor, \rightarrow, \neg\}$ at random with probabilities $q$. \;
\If{$\mathit{op} = \neg$}{
    $B := \mathit{self}(n-1)$.  \Comment*{Recursive call}
    \Return{$\neg B$}
}
\Else{
    $x$ := Choose an integer between 1 and $n-2$, inclusive, uniformly at random.\;
    $B := \mathit{self}(x)$. \;
    $C := \mathit{self}(n-1-x)$. \;
    \Return{$B\mathbin{\mathit{op}}C$}.
}

\caption{Random formula generation algorithm. \label{alg:formula}}
\end{algorithm}

\subsection{Implementation Detail}
We describe the details of the implementation during our experiments.

We represent proofs and sequents in IPL as the terms in simply-typed lambda calculus, implemented in the Rust programming language.
We verified the correctness of our proof by generating proof terms that can be read in Coq's interpreter and performing type checking.

We trained neural networks using the deep learning framework Chainer~\cite{chainer_learningsys2015}.
We used the implementations from the Chainer Chemistry library \footnote{https://github.com/pfnet-research/chainer-chemistry} for GNNs.

We performed experiments on Ubuntu~16.04 with a single CPU (Intel Core i7-4790, 3.60GHz).
We used a single GPU (NVIDIA Tesla K40c) for the inference of neural networks.

Throughout the experiment, we used $\gamma=0.95$ as a discount-rate, which we determined by a grid-search for proving as many theorems in the training library as possible.
We used $N_{\ge 2}=1000$ and $N_{=1}=100$ during data augmentation.

\subsubsection{Organization of Theorem Library}
Since we could not find a suitable publicly available theorem library in IPL for our setting~\footnote{For example, the ILTP library \texttt{http://www.iltp.de} contains only 128 provable theorems, which may not be sufficient for training neural networks even when we use our data augmentation procedure.},
we generated formulas using a random generation algorithm (Algorithm~\ref{alg:formula}).
Given a ``desired'' length parameter $n$, we recursively generate a syntax tree from top to bottom.
When we generate a leaf node, we uniformly choose a propositional variable from candidates at random.
When we generate an intermediate node, we randomly choose a logical symbol.
A generated formula has a length of between $n/2$ and $n$.

Note that the organization of our theorem library (the random generation) is entirely independent of our method.
Our method is applicable for any theorem library, even human-made ones.

In our experiment, we generated $D_\mathrm{train}$ with desired length $n \in [50, 400]$ and number of propositional variables $m \in [2, 20]$.
We generated $D_\mathrm{exam}$ with $n=500$ and $m \in [2,20]$.
The parameters here were uniformly sampled at random.
We sampled the probability vector $q$ from a Dirichlet distribution with $\alpha_j=3$ ($j \in [1,4]$), where $\alpha_j$ is the parameter of the Dirichlet distribution.
Since generated formulas are not necessarily provable in IPL, we checked their validity using an existing IPL prover.
We generated 2000 IPL theorems for $D_\mathrm{train}$ and 1000 for $D_\mathrm{exam}$.
The lengths of formulas of $D_\mathrm{train}$ were between 25 and 400, and those in $D_\mathrm{exam}$ were between 250 and 500.

\subsubsection{Rival Architectures}

We provide the details of the network architectures we compared against GNN in our prediction task.

\begin{description}
    \item[Bag-of-words]
        For a given sequent $p$, let $v_a(p)$ be the vector that records the counts of each propositional variable(i.e. $P_1, P_2$) and symbol (e.g, ($\land$, $\lor$) in the antecedants of $p$ .  Also let $v_c(p)$ be analogously defined vector for the  consequent of $p$. 
        We can then concatenate $v_a(p)$ and $v_c(p)$ to construct a vector whose dimension is two times the number of propositional variables and symbols present in the system.  Our \textit{Bag-of-words} predictor is a linear function of the form $\theta^T v(p)$.
    \item[Recurrent neural network]
        Recurrent neural networks (RNNs) are widely used architecture that can be applied to data of variable-length, such as natural languages.
        Since logical formula can be represented as a string, we can use RNNs for function approximation.
        In this paper, we experimented with two variants of RNNs: GRU \cite{chung2014empirical} and LSTM \cite{hochreiter1997long}.
\end{description}

\subsubsection{Hyperparameters}
To train neural networks, we used the Adam optimizer with learning rate $0.001$ and batch size~32.
We set the embedding size of RNNs to 32 and the dimension of hidden layers to 150.
When we used RNNs, we converted a sequent to a string in reverse Polish notation, which worked slightly better than a standard infix notation.
We set the unit size of the GNN to 16 and the step size $T$ of the GNN to 6.
After the application of $T$ embedding layers in the GNN, the hidden embeddings were summed up and fed to a multi-layer perceptron with one hidden layer of unit size 16.
These parameters were chosen to minimize validation loss.

When we benchmarked the inference time, we ran the inference with batch size~4 for the test set $D_0$.
This was an average batch size for computing Equation~\ref{eq:def-pi} when we ran the greedy DFS for training data.

During the prediction task, we trained each neural network for 20~epochs in a single machine.

We trained GNNs for 10~epochs during the proving task.
We ran the data augmentation procedure using 30~cluster machines to accelerate the algorithm.
Note that we only used cluster machines during approximate policy iteration.
We used a single machine for all other tasks, such as the evaluations in Tables~\ref{table:policy-iteration}~and~\ref{table:coq}.

\subsection{Analysis on Experiments}

\begin{figure}[t!]
  \centering
  \includegraphics[width=8cm]{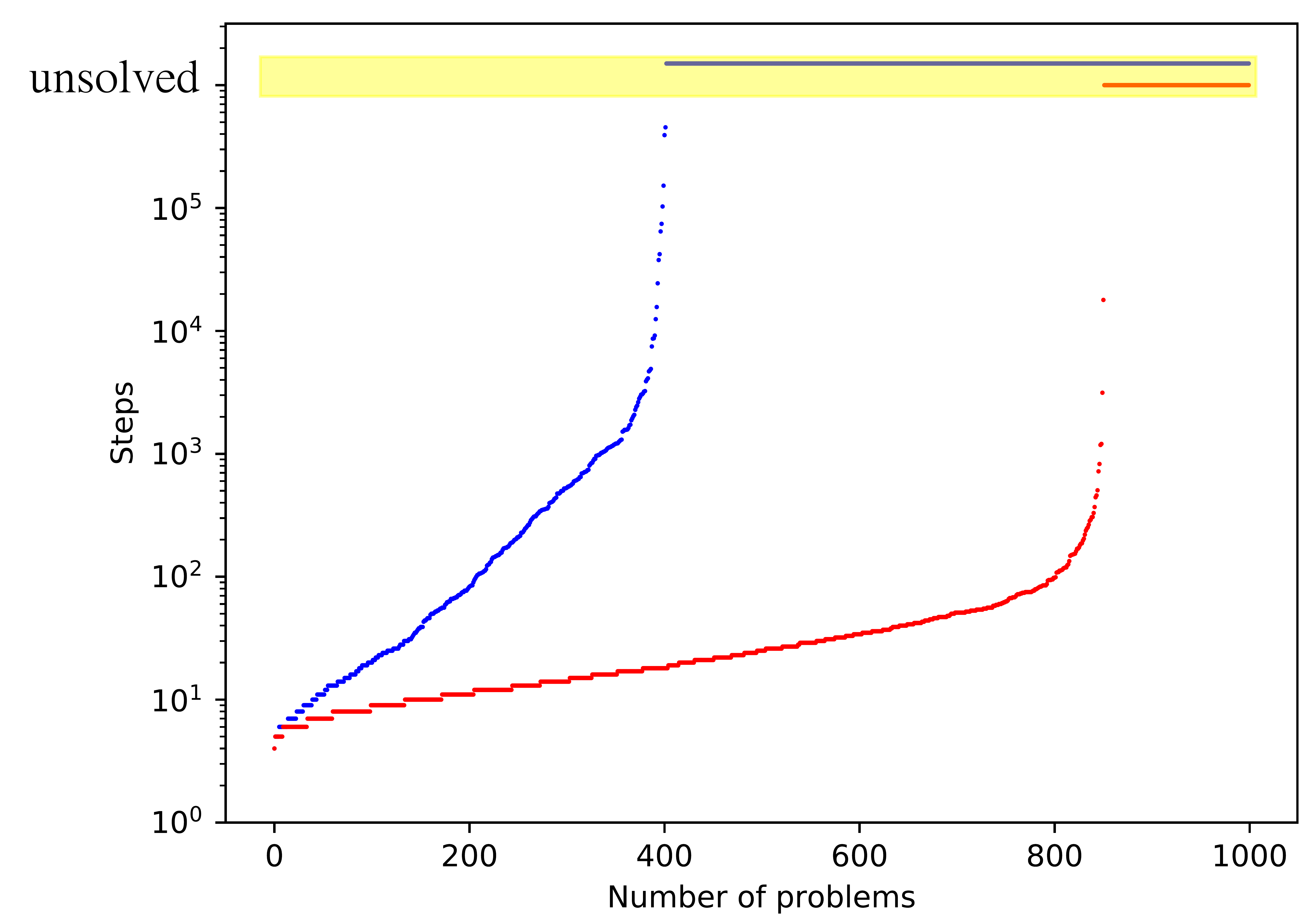}
  \caption{A cactus plot showing the number of steps during the greedy DFS. Red is the decisions by $\pi_4$, and blue is by $\pi_0$.
  \label{fig:cactus0}}
\end{figure}

\begin{figure}[t!]
  \centering
  \includegraphics[width=8cm]{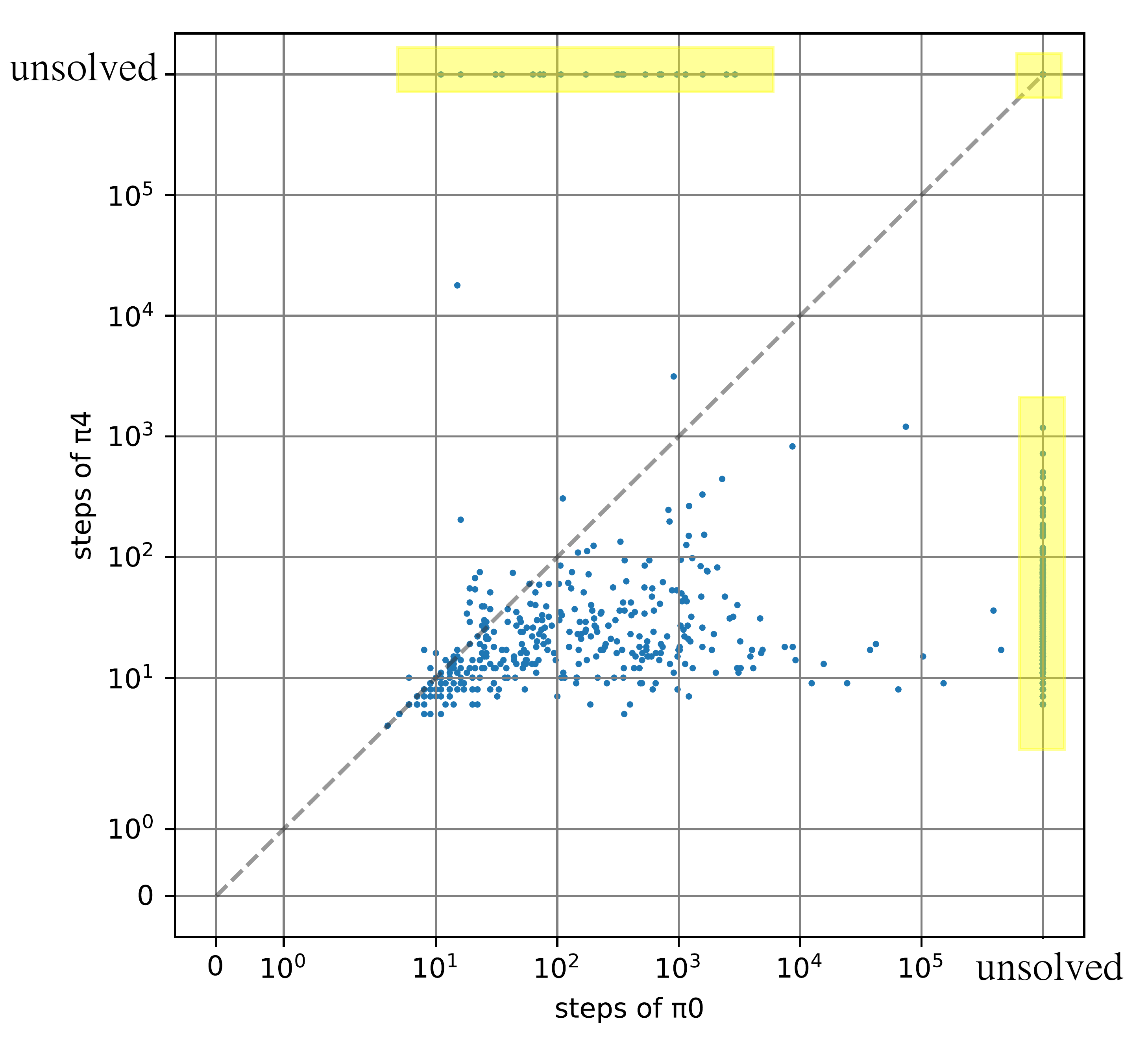}
  \caption{A scatter plot of the number of steps during the greedy DFS.
  \label{fig:cactus2}}
\end{figure}

In the proving task, we further examined how our prover could reduce the number of steps (explored states) during the search.
We compared the number of steps made by our baseline prover $\pi_0$ and our trained prover $\pi_4$.
We ran the greedy DFS with a time limit of 10 seconds for the dataset $D_\mathrm{exam}$.
Note that we did not compare against Coq's \texttt{tauto} since the number of steps is unavailable from Coq's interface.

We show the result in Figures~\ref{fig:cactus0}~and~\ref{fig:cactus2}.
In Figure~\ref{fig:cactus0}, the horizontal axis corresponds to the number of solved problems, and the vertical axis corresponds to the number of steps.
The red curve corresponds to $\pi_4$ and the blue to $\pi_0$.
Each of these plots is sorted by the number of steps.
In Figure~\ref{fig:cactus2}, a single point corresponds to a problem in $D_\mathrm{exam}$.
The vertical axis is the number of steps by $\pi_4$, and the horizontal axis is by $\pi_0$.

The figures indicate that our trained prover $\pi_4$ roughly reduced the number of steps by a factor of several tens, depending on the problem.

\end{document}